%% file: main.tex
\pdfoutput=1
\documentclass[letterpaper,11pt]{article}

\usepackage[utf8]{inputenc}
\usepackage[T1]{fontenc}
\usepackage{times}
\usepackage[margin=1in,top=0.45in]{geometry}
\usepackage{microtype}

\usepackage{amsmath,amssymb,amsfonts,amsthm}
\usepackage{mathrsfs}
\usepackage{dsfont}
\usepackage{nicefrac}

\usepackage{graphicx}
\usepackage{tabularx}
\usepackage{longtable}

\usepackage{blindtext}
\usepackage{etoc}

\PassOptionsToPackage{numbers, sort&compress}{natbib}
\usepackage{natbib}
\usepackage{url}

\input{math_commands.tex}

\input{preamble.tex}
\usepackage{fontawesome5}

\theoremstyle{definition}

\renewcommand{\thefootnote}{\fnsymbol{footnote}}

\setlist{leftmargin=5mm}

\definecolor{abstractbg}{RGB}{230,242,250}
\definecolor{abstractborder}{RGB}{200,210,220}
\usepackage{titlesec}

\titleformat{\section}
  {\sffamily\bfseries\Large}
  {\thesection}
  {1em}
  {}

\renewenvironment{abstract}
{
\begin{center}
\begin{tcolorbox}[
    colback=abstractbg,
    colframe=abstractborder,
    boxrule=0.6pt,
    arc=6pt,
    width=\textwidth,
    left=8pt,
    right=8pt,
    top=8pt,
    bottom=8pt
]
}
{
\end{tcolorbox}
\end{center}
}

\newcommand{\papertitle}{%
\sffamily\bfseries\fontsize{16}{1}\selectfont
From Knowledge Access to Source Learning: \\[.35em] Developing Source-Specific Competence
}

\newcommand{\paperauthors}{%
\sffamily
Lucheng Fu$^{1}$, Kejing Xia$^{1}$, Yiyang Wang$^{1}$, Yiqiao Jin$^{1}$, Jinjin He$^{1}$,
\\[0.35em]
Xiyuan Yang$^{2}$, Haoxin Liu$^{1}$, Ye Yu$^{2}$, Haibo Jin$^{2}$, Yijia Xiao$^{3}$,
\\[0.35em]
Wenke Lee$^{1\dagger}$, B.~Aditya Prakash$^{1\dagger}$, Haohan Wang$^{2\dagger}$
\\[0.55em]
{\normalsize $^{1}$Georgia Institute of Technology \qquad $^{2}$University of Illinois at Urbana-Champaign}
\\[0.15em]
{\normalsize $^{3}$University of California, Los Angeles}
\\[0.45em]
{\normalsize \faGlobe~Website: \url{https://sourcelearn.github.io/}}
\\[0.1em]
{\normalsize \faGithub~GitHub: \url{https://github.com/luchengfu6/SourceLearn}}
\footnotetext[2]{Corresponding authors.}
{\renewcommand{\thefootnote}{}\footnotetext{Contact: \texttt{luchengfu@gatech.edu}}}
}

\newcommand{\paperdate}{}

\begin{document}

\thispagestyle{empty}

\noindent
\includegraphics[height=.8cm]{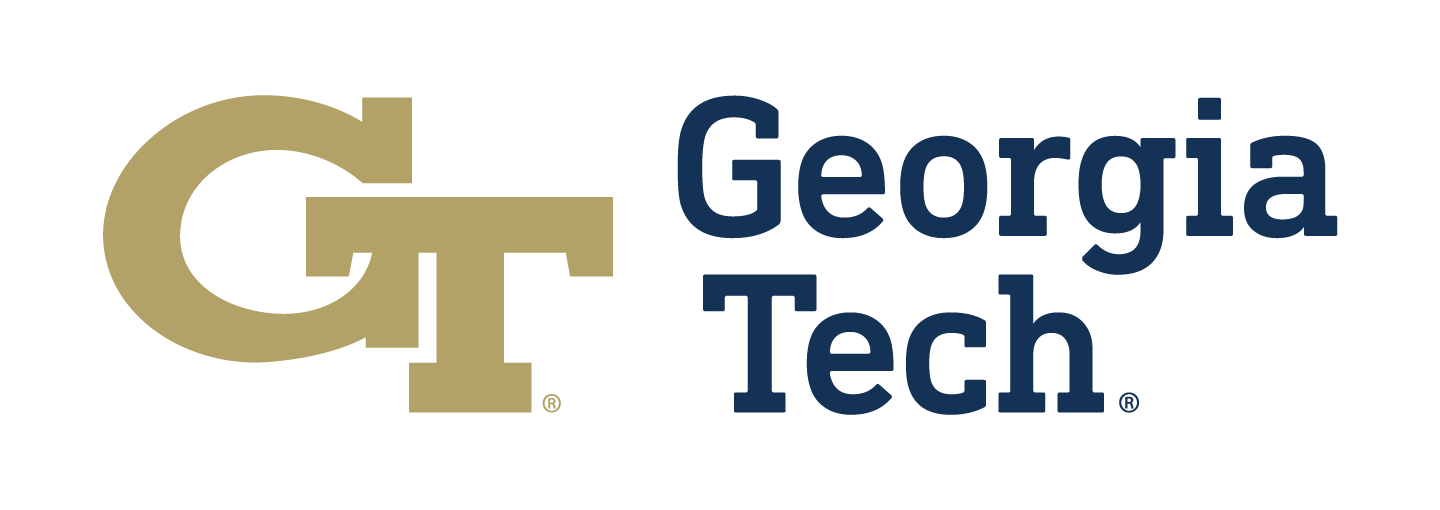}
\includegraphics[height=.8cm]{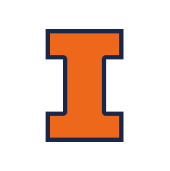}
\includegraphics[height=.8cm]{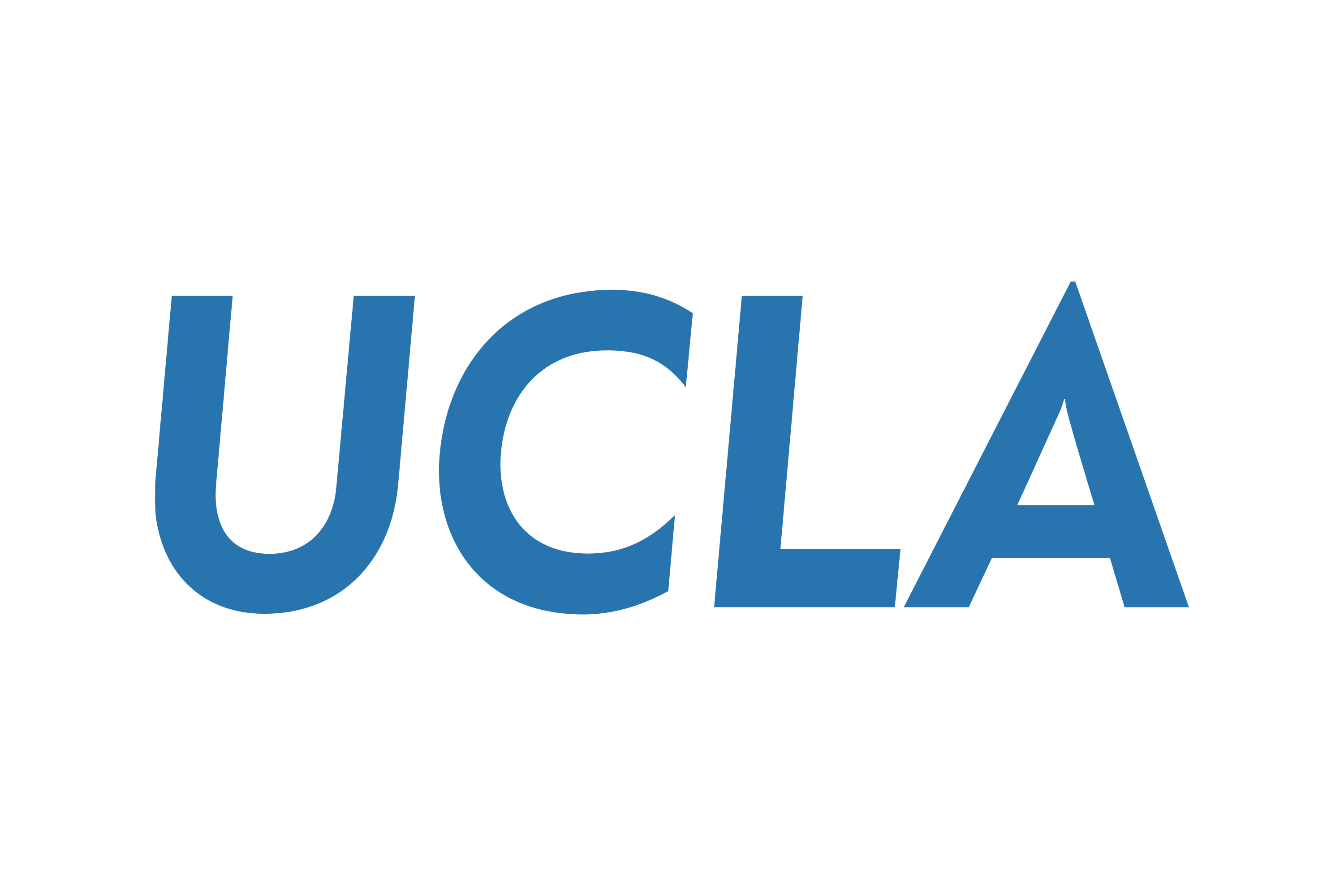}

\vspace{-.15in}

\noindent\rule{\textwidth}{0.8pt}

\vspace{0.65cm}

\begin{center}
    {\papertitle\par}
    \vspace{0.45cm}
    {\large \paperauthors\par}
    \vspace{0.2cm}
    {\normalsize \paperdate\par}
\end{center}

\begin{abstract}
\input{0_abstract}
\end{abstract}

\setcounter{footnote}{0}
\renewcommand{\thefootnote}{\arabic{footnote}}

\section{Introduction}
\label{sec:intro}
\input{1_introduction}

\section{Related Work}
\label{sec:related}
\input{2_related_work}

\section{Problem Formulation}
\label{sec:problem}
\input{3_Problem}

\section{Method}
\label{sec:method}
\input{4_method}

\section{Experiments}
\label{sec:exp}
\input{5_exp}

\section{Conclusion}
\label{sec:conclusion}
\input{6_conclusion}

\bibliographystyle{unsrt}
\bibliography{8_reference}

\newpage
\appendix
\input{7_appendix}

\end{document}

%% file: math_commands.tex
\usepackage{amsmath,amsfonts,bm}

\def\eqref#1{equation~\ref{#1}}

\def\1{\bm{1}}

\DeclareMathAlphabet{\mathsfit}{\encodingdefault}{\sfdefault}{m}{sl}
\SetMathAlphabet{\mathsfit}{bold}{\encodingdefault}{\sfdefault}{bx}{n}

%% file: preamble.tex
\newcommand{\method}{{\fontfamily{lmtt}\selectfont \textbf{SourceLearn}}\xspace}

\usepackage[textsize=tiny]{todonotes}
\usepackage[algo2e]{algorithm2e}
\usepackage[export]{adjustbox}
\usepackage{threeparttable}
\usepackage{multirow}
\usepackage{enumitem}
\usepackage{newfloat}
\usepackage{listings}
\usepackage{colortbl}
\usepackage{amsfonts}
\usepackage{amssymb}
\usepackage{pifont}
 \usepackage{booktabs}
\usepackage{float}
\usepackage{bm}
\usepackage{tabulary}
\usepackage{makecell}
\usepackage{bbm}
\usepackage{mdframed}
\usepackage{algorithm}
\usepackage{algorithmic}
\usepackage{wrapfig}
\usepackage{caption,subcaption}
\makeatletter
\@ifpackageloaded{hyperref}{}{\usepackage[pagebackref=false,breaklinks=true,colorlinks,bookmarks=false]{hyperref}}
\makeatother
\usepackage[most]{tcolorbox}
\tcbuselibrary{theorems}
\definecolor{lightgray}{gray}{.9}
\definecolor{deepgray}{gray}{.8}
\tcbset{highlight math/.append style={left=0mm,right=0mm,top=0mm,bottom=0mm, colframe=white}}

\usepackage[capitalize,noabbrev]{cleveref}

\newcolumntype{I}{!{\vrule width 1pt}}
\makeatletter
\newcommand{\thickhline}{%
    \noalign {\ifnum 0=`}\fi \hrule height 1pt
    \futurelet \reserved@a \@xhline
}
\makeatother

\crefname{proposition}{Prop.}{Props.}
\crefname{section}{Sec.}{Secs.}
\crefname{table}{Tab.}{Tabs.}

\usepackage{xspace}
\makeatletter
\DeclareRobustCommand\onedot{\futurelet\@let@token\@onedot}
\def\@onedot{\ifx\@let@token.\else.\null\fi\xspace}

\usepackage[dvipsnames]{xcolor}
\usepackage{colortbl}
\definecolor{ada_blue}{rgb}{0,205,205}
\definecolor{glt_red}{rgb}{109,205,255}
\definecolor{MorandiBlue}{RGB}{118,134,146}
\definecolor{demphcolor}{RGB}{144,144,144}
\definecolor{mygray}{gray}{0.4}
\definecolor{autopurple}{HTML}{7030A0}
\definecolor{dyna_yellow}{HTML}{BF9000}
\definecolor{adaptive_blue}{HTML}{0070C0}
\definecolor{darkgrey}{RGB}{120,120,120}
\definecolor{mygrey}{RGB}{200,200,200}
\definecolor{myblue}{HTML}{00CDCD}
\definecolor{champagne}{rgb}{0.97, 0.91, 0.81}
\definecolor{darksalmon}{rgb}{0.91, 0.59, 0.48}
\definecolor{emerald}{rgb}{0.31, 0.78, 0.47}
\definecolor{green(pigment)}{rgb}{0.0, 0.65, 0.31}
\definecolor{amaranth}{rgb}{0.9, 0.17, 0.31}
\definecolor{iris}{rgb}{0.35, 0.31, 0.81}
\definecolor{uu}{rgb}{0.95, 0.51, 0.51}
\definecolor{spirodiscoball}{rgb}{0.06, 0.75, 0.99}
\definecolor{cadetblue}{RGB}{95,158,160}
\definecolor{keywordcolor}{RGB}{178,34,34}

\definecolor{customgreen}{HTML}{667b5b}

\definecolor{customblue}{HTML}{bcccea}

\definecolor{flcyellow}{RGB}{180,140,0}

\usepackage{pifont}
\usepackage{bbding}

\SetCommentSty{mycommfont}

\newcommand{\reddown}[1]{_{\color{darksalmon}\downarrow #1}}
\newcommand{\greenup}[1]{_{\color{green(pigment)}\uparrow #1}}

\crefname{equation}{Eq.}{Eqs.}
\Crefname{equation}{Equation}{Equations}

%% file: 0_abstract.tex
Large language model (LLM) agents increasingly rely on persistent external sources to solve sequences of knowledge-intensive tasks. Existing methods improve how source content is accessed and organized, while agent-memory systems preserve reusable knowledge from prior interactions, but repeated use of the same source is still largely treated as repeated access rather than an opportunity to progressively improve understanding of that source. We study \emph{source learning}: developing reusable \emph{source-specific competence} over a persistent authoritative source. We represent this competence with a persistent \emph{source model} that captures reusable understanding of the source, including how its knowledge is structured, interpreted, and applied. To construct and progressively refine such models, we propose \method{}, which combines two complementary learning mechanisms. \emph{Self-Directed Source Learning} identifies what remains incompletely understood and adaptively revisits the source, while \emph{Task-Guided Source Learning} uses downstream experience to reveal local representational gaps and recurring needs in how source knowledge should be organized. In both cases, learning signals determine what should be reconsidered, while persistent updates are reconstructed from the authoritative source. Across five benchmarks and three LLM backends, \method{} achieves the best performance in 13 of 15 settings, with gains of up to 22.6 points over Hybrid RAG and substantial overall improvements over static source representations and experience-based memory baselines.

%% file: 1_introduction.tex
Large language model (LLM) agents increasingly rely on external sources to solve knowledge-intensive tasks. Here, a \emph{source} denotes an identifiable body of external knowledge, such as a document collection, a code repository, or API documentation.
Existing work has substantially improved how such sources are accessed and organized through retrieval, long-context modeling, and structured source representations~\citep{lewis2020retrieval,xu2024retrieval,sarthi2024raptor,edge2024local,gutierrez2024hipporag,gutierrez2025rag}.
In parallel, agent-memory methods preserve reusable knowledge acquired from prior interactions so that experience can influence future behavior~\citep{shinn2023reflexion,zhao2024expel,xu2026mem}.
These directions make source content easier to access and organize, and past experience easier to reuse, but they optimize different objects: source-access methods primarily support query-time use of external knowledge, while agent memory primarily preserves knowledge derived from previous interactions.
When many tasks depend on the same source, an additional opportunity arises: the agent can use prior study and task experience to improve how it represents and uses that source.
This motivates a complementary learning problem: can repeated interaction with the source progressively develop reusable competence about the source itself, beyond repeatedly accessing its content or simply accumulating task experience?

We call this process \emph{source learning}. Its goal is to develop \emph{source-specific competence}: reusable understanding of a source that can be applied to new tasks grounded in that source. Such competence goes beyond retaining or compressing source content: it captures how source knowledge is structured, interpreted, and applied, including relationships among concepts, conditions governing their applicability, procedures, and how different pieces of information combine to support future reasoning and action. For example, an agent studying API documentation may learn how prerequisites and dependencies connect multiple calls, enabling it to reason about new combinations of those calls. In our setting, the source is \emph{persistent} in that it remains available across tasks, and \emph{authoritative} in that it serves as the reference for factual content. The agent's accumulated understanding complements rather than replaces direct access to this source. Repeated study and use can reveal gaps in this understanding and guide its refinement, turning source use into a cumulative learning process rather than a sequence of isolated access events.

\input{figure/fig-problem}
Developing such competence raises two key challenges. \hypertarget{Q1}{\textbf{\uppercase\expandafter{\romannumeral1})}} \textbf{\textit{What source knowledge should be learned before future tasks are known?}} A source may contain far more information than should be retained, while a one-pass representation may fail to connect information distributed across the source or capture conditions that are only implicit.
Therefore, source learning must identify which knowledge is broadly reusable and organize it into a form that can support future tasks whose exact demands are not yet known.
\hypertarget{Q2}{\textbf{\uppercase\expandafter{\romannumeral2})}} \textbf{\textit{How can task-specific feedback improve source-specific competence beyond the task that produced it?}} Downstream use may expose deficiencies that prior study did not anticipate, but directly retaining the observed answer or correction risks learning the task rather than the source.
Effective adaptation must instead use task-level evidence to identify deficiencies in source understanding and address them with reusable, source-grounded knowledge.

We introduce \method{}, which maintains a persistent \emph{source model} through two complementary learning mechanisms. \emph{Self-Directed Source Learning} revisits incompletely understood source content and connects knowledge across entities. \emph{Task-Guided Source Learning} uses failures to identify local representational gaps and recurring task demands to improve how knowledge is organized. Both mechanisms use observations to guide learning while reconstructing persistent updates from the authoritative source.

Our contributions are threefold:
\begin{itemize}[leftmargin=*,itemsep=2pt,parsep=0pt,topsep=2pt]
\item[\ding{182}] \textbf{Source Learning Formulation and Representation.} We formulate \emph{source learning} as the development of reusable \emph{source-specific competence} and introduce the \emph{source model} as an explicit, persistent, and revisable representation of that competence, complementing direct source access while preserving the source as the factual authority.

\item[\ding{183}] \textbf{Self-Directed and Task-Guided Source Learning.} We develop two complementary mechanisms for progressively building source-specific competence: Self-Directed Source Learning proactively studies the source to deepen incomplete understanding and connect fragmented knowledge, while Task-Guided Source Learning uses deficiencies and recurring demands revealed through downstream tasks to further refine that understanding.

\item[\ding{184}] \textbf{Strong Empirical Performance.} We evaluate \method{} across document QA, code QA, tool use, and interactive environments. It substantially improves overall same-source task performance over task-local retrieval, static source representations, and experience-based memory baselines, with gains from both self-directed and task-guided source learning.
\end{itemize}

%% file: figure/fig-problem.tex
\definecolor{P1}{RGB}{191, 144, 0}
\definecolor{P2}{RGB}{14, 132, 140}

\begin{wrapfigure}{r}{0.5\textwidth}
    \vspace{-15pt}
    \begin{center}
        \includegraphics[width=0.5\textwidth]{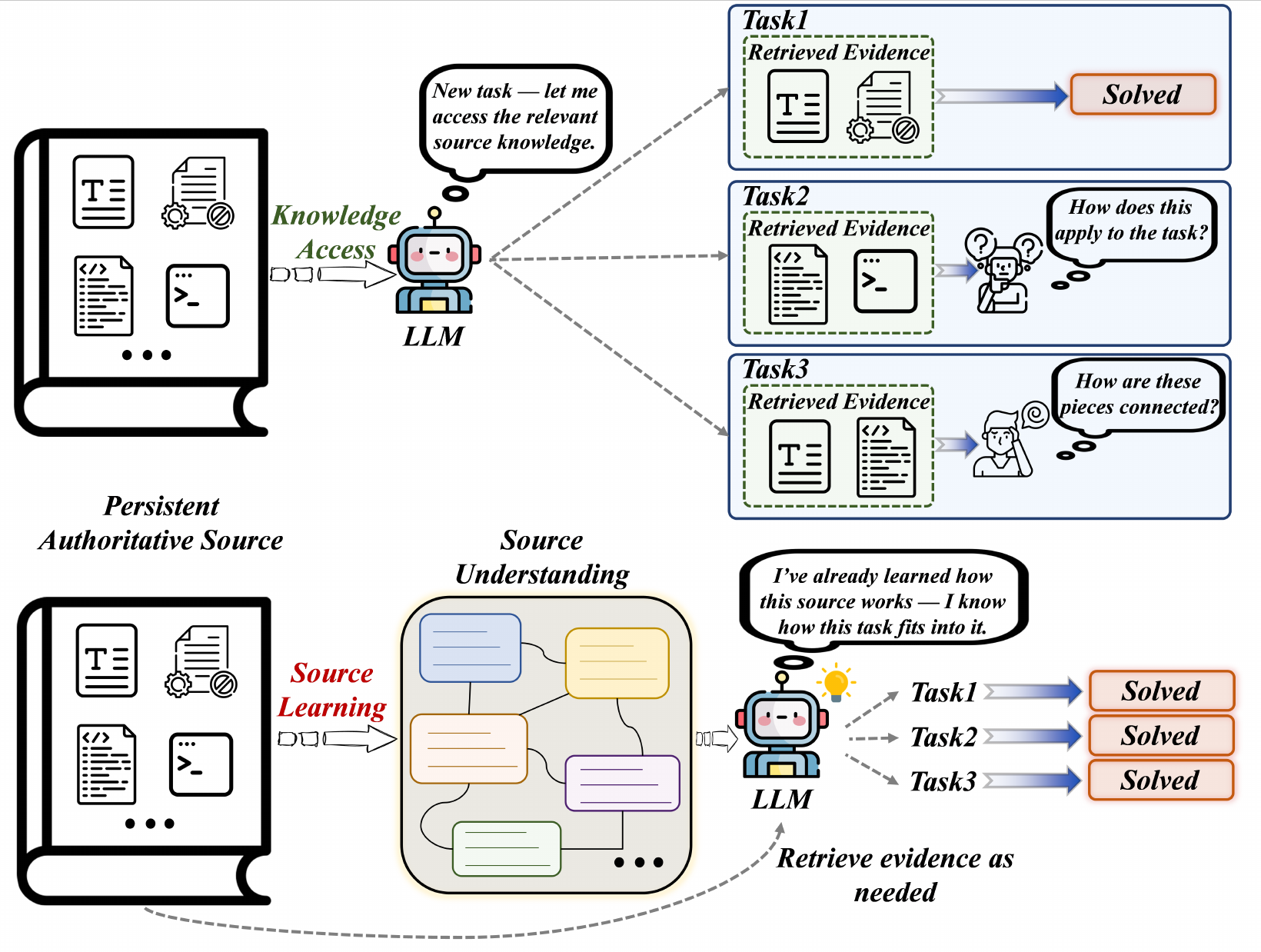}
    \end{center}
    \vspace{-15pt}
    \caption{\small
    \textbf{Problem Illustration.}
    \textbf{\uppercase\expandafter{\romannumeral1})}
    \textcolor{P1}{Knowledge access} retrieves task-relevant evidence from the source, but does not necessarily accumulate reusable source understanding across tasks, requiring the model to repeatedly reconstruct how evidence should be interpreted or combined.
    \textbf{\uppercase\expandafter{\romannumeral2})}
    Our goal is \textcolor{P2}{source learning}: progressively developing reusable source understanding through continued study and use, while retaining direct access to the authoritative source.}
    \label{fig:problem}
    \vspace{-15pt}
\end{wrapfigure}

%% file: 2_related_work.tex
\subsection{External Knowledge Access and Source Representations}

Retrieval-augmented language models incorporate external knowledge through task-time retrieval or retrieval-augmented pretraining~\citep{guu2020retrieval,karpukhin2020dense,lewis2020retrieval,izacard2021leveraging,borgeaud2022improving,izacard2023atlas,shi2024replug}. Subsequent work improves when and how evidence is accessed through iterative, active, adaptive, or self-reflective retrieval~\citep{trivedi2023interleaving,jiang2023active,asai2024self,jeong2024adaptive,yan2024corrective}, while long-context methods expose larger portions of the source directly but still face challenges in reliably using distributed information~\citep{xu2024retrieval,liu2024lost}. Complementary approaches structure source knowledge: RAPTOR builds hierarchical summaries~\citep{sarthi2024raptor}; GraphRAG, LightRAG, GraphReader, and KAG use graph-based structures~\citep{edge2024local,guo2025lightrag,li2024graphreader,liang2025kag}; and HippoRAG, HippoRAG 2, and MemoRAG support associative retrieval, continual knowledge integration, and global memory, respectively~\citep{gutierrez2024hipporag,gutierrez2025rag,qian2025memorag}. StructRAG constructs task-conditioned representations at inference time~\citep{li2025structrag}. These approaches primarily improve how source knowledge is accessed, organized, or retained. In contrast, \method{} treats competence over a persistent authoritative source as the object of learning itself, progressively refining a reusable source model through continued study and use.

\subsection{Agent Memory and Experiential Learning}

Agent-memory methods enable LLM agents to retain and reuse information accumulated through prior interactions~\citep{wang2024survey,zhang2025survey}. Existing systems maintain conversational, episodic, or hierarchical memories~\citep{park2023generative,zhong2024memorybank,liu2023think,packer2023memgpt}, or learn from task experience by retaining feedback, abstracting reusable insights, and acquiring skills or workflows~\citep{shinn2023reflexion,zhao2024expel,wang2024voyager,wang2024agent}. More recent work further organizes such experience into structured and revisable representations, including knowledge-graph world models in AriGraph~\citep{anokhin2024arigraph}, interconnected memories in A-MEM~\citep{xu2026mem}, and procedural memory in ReMe~\citep{cao2026remember}. \method{} differs in the object and authority of learning: task experience is used to reveal deficiencies or recurring representational needs in source understanding, while persistent updates are reconstructed from the authoritative source rather than stored directly from interaction experience.

%% file: 3_Problem.tex
\subsection{Knowledge Access over a Persistent Source}

We consider a persistent authoritative source $D$ and a sequence of tasks $(q,y)\sim\mathcal{P}_D$ grounded in the same source. Let $R(D,q)$ denote the source context made available for task $q$, and let $F$ denote the task-solving LLM agent. Existing source-grounded systems can be broadly written as
\begin{equation}
    \hat{y}=F\big(q,R(D,q)\big),
    \label{eq:knowledge-access}
\end{equation}
where $R$ may represent retrieval, long-context access, or a structured source representation. These methods differ in how source knowledge is organized and accessed, but primarily aim to provide useful information for the current task.

\paragraph{Access Does Not Imply Source Learning.}
Repeated access to the same source does not necessarily improve the system's understanding of that source. Even with persistent indexes or structured representations, reusable knowledge may remain fragmented or need to be reconstructed across tasks. Source learning instead asks whether repeated study and use can progressively develop reusable understanding that carries across future tasks.

\subsection{Source-Specific Competence}
\label{sec:source-competence}

We define \emph{source-specific competence} as reusable understanding of a source that can support new tasks grounded in that source, including how its knowledge is structured, interpreted, and applied. We represent this competence with a persistent and revisable \emph{source model} $M$. Given the source $D$ and accumulated study and task experience $H_t$, source learning develops
\begin{equation}
    M_t=\operatorname{Learn}(D,H_t),
    \label{eq:source-learning}
\end{equation}
so that understanding acquired through earlier study and use can be carried forward to future tasks.

For a future task, this accumulated understanding complements rather than replaces direct source access:
\begin{equation}
    \hat{y}=F\big(q,R(D,q),M_t\big),
    \label{eq:competence-answer}
\end{equation}
where $R(D,q)$ provides task-specific source information, while $M_t$ provides reusable understanding accumulated through prior study and use. The original source remains authoritative for factual content. Source learning therefore seeks to improve future same-source task performance by progressively developing stronger and more reusable understanding of the source.

%% file: 4_method.tex
\subsection{Overview}
\label{sec:method-overview}

\method{} develops reusable competence over a persistent authoritative source $D$ by maintaining an explicit and revisable source model $M$. Rather than treating each task as an independent access event, the system carries forward an organized understanding of the same source and progressively develops that understanding through continued study and use.

The source model evolves through two complementary learning mechanisms. \emph{Self-Directed Source Learning} uses the current model to guide renewed study of the source, with the goal of deepening incomplete understanding and connecting knowledge that remains fragmented. \emph{Task-Guided Source Learning} uses downstream task experience to reveal both concrete representational deficiencies and recurring preferences about how source knowledge should be represented. The overall process is
\begin{equation}
M_0 = \operatorname{Construct}(D), \quad M_S = \operatorname{SelfLearn}(D,M_0), \quad M_T = \operatorname{TaskLearn}(D,M_S;\mathcal{G}).
\label{eq:overall-learning}
\end{equation}
Here, $M_0$ is the initial source model, $M_S$ is the model after self-directed learning, and $M_T$ is the final model after task-guided learning. $\mathcal{G}=\{(q_t,y_t)\}_{t=1}^{T}$ denotes a set of guidance tasks grounded in the same source.

\input{figure/fig-overview}

\subsection{Source Model Construction and Activation}
\label{sec:source-model}

\paragraph{Source representation.}
We organize the source model around entities $e$, each associated with a reusable representation $m_e$. Entity representations capture source-grounded understanding that may be useful across tasks, such as important properties, mechanisms, governing conditions, procedures, exceptions, distinctions, and relations to other entities. Beyond individual entities, $M$ may also contain higher-level structure that organizes recurring patterns and dependencies across multiple entities.

The purpose of $M$ is not to reproduce all of $D$. Instead, it maintains a compact and reusable understanding of the source while leaving easily retrievable low-level details in the original source. This separation is important: the source model captures what prior study has learned about the source, while $D$ remains directly available as the authoritative factual reference.

\paragraph{Initial construction.}
We construct a provisional source model $M_0$ through source-aware, entity-centered reading. Large sources may first be partitioned into coherent regions, while native organization such as code modules, symbols, or document sections is used when available. Within each region, the reader constructs representations of the major entities and organizes reusable structure that spans them.

Construction follows a generic modeling instruction $\Pi_0$ that favors major entities, important relations, procedures, governing conditions, distinctions, and representative details while avoiding redundant low-level content. The resulting $M_0$ is deliberately provisional: it provides a structured starting point for subsequent learning rather than an exhaustive representation of the source.

\paragraph{Grounded source reconstruction.}
All later learning stages follow a common principle: learning signals determine \emph{what should be reconsidered}, while persistent knowledge must be reconstructed from the authoritative source. For an affected source region $R$,
\begin{equation}
O_R = \operatorname{Inspect}(D_R,M_R;\xi), \quad
M' = \operatorname{GroundApply}\big(M,\operatorname{Reconstruct}(M_R,O_R);D_R\big).
\label{eq:grounded-reconstruction}
\end{equation}
Here, $\xi$ is the learning signal and $O_R$ contains temporary observations obtained by rereading the relevant source content against the current model.

A reconstruction is committed only when the proposed content is supported by source evidence and previously supported meaning is preserved unless it is subsumed by a stronger representation. Reconstruction may merge fragmented knowledge, make important conditions or relations explicit, or remove redundant organization, but observations themselves are never copied directly into persistent memory. This shared reconstruction mechanism is used by both learning stages below.

\paragraph{Source-model activation.}
The source model may eventually exceed the context budget available to a task. We therefore expose at most $B$ tokens of accumulated source understanding:
\begin{equation}
A_B(M,q)=
\begin{cases}
M, & \operatorname{tok}(M)\leq B,\\
\operatorname{ActivateRegions}(M,q;B), & \operatorname{tok}(M)>B.
\end{cases}
\label{eq:budgeted-activation}
\end{equation}
When activation is necessary, it operates over coherent source-model regions established during construction, selecting task-relevant regions while preserving their internal structure.

For a task $q$, the solver receives both activated source understanding and task-specific evidence directly retrieved from the authoritative source:
\begin{equation}
\hat{y}=F\big(q,R(D,q),A_B(M,q)\big).
\label{eq:task-time-use}
\end{equation}
The two contexts therefore serve complementary roles: $A_B(M,q)$ provides reusable understanding accumulated through prior study and use, while $R(D,q)$ preserves direct access to precise source evidence.

\subsection{Self-Directed Source Learning}
\label{sec:self-learning}

An initial source model cannot anticipate everything that may later prove useful to understand. Some mechanisms may remain only partially represented, governing conditions may remain implicit, and knowledge distributed across multiple entities may remain fragmented. Self-Directed Source Learning therefore uses the current model to guide renewed study of the source, allowing the system to decide what deserves deeper investigation before future task requirements are known.

Each self-study cycle follows an \emph{Inspect--Study--Consolidate} process:
\begin{equation}
O = \operatorname{Inspect}(D,M), \quad O^{+} = \operatorname{Study}(D,M,O), \quad M' = \operatorname{Consolidate}(D,M,O^{+}).
\label{eq:self-directed-learning}
\end{equation}

\paragraph{Model-conditioned source inspection.}
The cycle begins with a broad inspection of the source conditioned on the current model. For each entity $e$, the system rereads its associated source content $S_e$ together with its current representation $m_e$:
\begin{equation}
O_e = \operatorname{Observe}(S_e,m_e).
\label{eq:self-inspection}
\end{equation}
Rather than summarizing $S_e$ again, inspection asks what the current representation still fails to explain well. For example, a mechanism may be represented only partially, the governing condition of a rule may remain implicit, or a reference to another entity may not yet be integrated into the model.

Each observation records the relevant source evidence and its implication for the current representation. These observations are provisional: they do not directly modify $M$, but instead identify promising directions for further study and provide material for later consolidation.

\paragraph{Adaptive source study.}
Inspection may reveal many possible gaps, not all of which are equally important. At planning round $b$, an adaptive planner selects a small set of study actions using the current model, accumulated observations, and the history of previous study:
\begin{equation}
\Gamma_b = \operatorname{PlanStudy}(M,\overline{O}_b,H_{<b}), \quad
\Gamma_b \subseteq \left\{
\textsc{Deepen}(e,q^{\mathrm{study}}),
\textsc{Connect}(e_i,e_j,q^{\mathrm{study}})
\right\}.
\label{eq:study-planning}
\end{equation}
\textsc{Deepen} investigates an unresolved aspect of one entity, such as a mechanism, condition, or procedure. \textsc{Connect} jointly studies two entities when inspection suggests that an important dependency or shared structure remains insufficiently understood.

For each action, the planner formulates a focused temporary study question $q^{\mathrm{study}}$, which guides retrieval and renewed source reading. The resulting evidence produces additional observations that can inform subsequent study rounds. The questions themselves are learning instruments and are not stored in the source model.

\paragraph{Evidence consolidation.}
Inspection and study accumulate observations while leaving the persistent model unchanged. Once study terminates, the accumulated observations are consolidated through the grounded reconstruction mechanism in \cref{eq:grounded-reconstruction}. Entity-specific observations revise the corresponding representations, while cross-entity observations can update higher-level relational structure.

This produces a \emph{read-many, write-once} pattern: the system can broadly inspect and selectively study the source before committing a persistent update. Applying this process to $M_0$ yields the self-directed source model $M_S$.

\subsection{Task-Guided Source Learning}
\label{sec:task-learning}

Self-directed learning asks what the current source model still needs to understand without assuming a particular future task. Actual downstream use provides a complementary signal: it reveals which parts of source knowledge matter in practice and whether the current representation makes them available in a useful form.

Task experience can reveal deficiencies at two levels. A particular failure may show that some required source knowledge is missing from the model, while repeated tasks may reveal a broader preference about \emph{how} source knowledge should be represented. We therefore perform Task-Guided Source Learning through two complementary pathways: \emph{failure-guided local refinement} and \emph{cross-task representation learning}.

\paragraph{Diagnosing source requirements from task use.}
For each guidance task $(q_t,y_t)\in\mathcal{G}$, we first return to the authoritative source to identify what source evidence supports the reference outcome and what source understanding the task requires:
\begin{equation}
\widetilde{E}_t = \operatorname{RetrieveGuide}(D,q_t,y_t), \quad (E_t^{*},C_t) = \operatorname{Diagnose}(q_t,y_t,\widetilde{E}_t).
\label{eq:task-evidence}
\end{equation}
Here, $E_t^{*}$ contains the source evidence that directly supports the reference outcome, while $C_t$ contains the corresponding source requirements. Thus, $E_t^{*}$ specifies where the required support comes from, whereas $C_t$ describes what understanding the source model would need to make available.

This is the only stage of source learning that directly observes $y_t$. The reference outcome is used only to locate authoritative support and derive source requirements; neither the answer nor its supporting evidence is stored directly as persistent knowledge. The guidance task itself is solved using the same task-time interface in \cref{eq:task-time-use}.

\paragraph{Failure-guided local refinement.}
If a guidance task is answered incorrectly, we determine which of its requirements are absent from the current source model:
\begin{equation}
C_t^{-}=\{c\in C_t:c\text{ is not represented in }M_t\}, \quad M_{t+1}=\operatorname{Refine}(D,M_t;C_t^{-}).
\label{eq:local-refinement}
\end{equation}
When $C_t^{-}\neq\varnothing$, the missing requirements identify the part of the source model whose current representation is insufficient. They act only as learning targets: the implicated source region is reread from $D$ and revised through grounded reconstruction rather than by storing the task correction itself.

Requirements that are already represented do not trigger additional writing, which separates a representational deficiency from an error that occurs despite the required source understanding already being available. Processing the guidance tasks in this way produces a locally refined model $M_{\mathrm{local}}$.

\paragraph{Cross-task representation learning.}
Task experience also provides a more transferable signal. Across multiple tasks, the system may repeatedly encounter cases where certain distinctions should remain explicit, conditions should not be collapsed, related knowledge should be represented together, or information should be maintained at a different level of granularity.

We summarize the signal from each guidance task as a representation lesson and aggregate recurring lessons into a source-level representation policy:
\begin{equation}
\ell_t = \operatorname{Lesson}(q_t,C_t,M_t,\hat{y}_t), \quad \Pi_1 = \operatorname{Aggregate}\big(\Pi_0,\{\ell_t\}_{t=1}^{T}\big).
\label{eq:representation-policy}
\end{equation}
Each lesson describes a preference about \emph{how} source knowledge should be represented rather than a particular source fact or task answer. Lessons are constrained to avoid task-specific identifiers, and $\Pi_1$ retains recurring preferences that can generalize beyond the task that produced them.

Importantly, $\Pi_1$ does not itself introduce new source knowledge. Instead, it indicates what kinds of source structure deserve clearer or more explicit representation. We therefore return to the authoritative source and recalibrate the locally refined model under the learned policy:
\begin{equation}
M_T = \operatorname{Recalibrate}(D,M_{\mathrm{local}};\Pi_1).
\label{eq:source-recalibration}
\end{equation}
Recalibration revisits the source under $\Pi_1$, identifies source-supported structure that is currently missing or represented at an inappropriate granularity, and incorporates it through the same grounded reconstruction mechanism.

Task-Guided Source Learning therefore transfers experience at two levels: concrete failures identify local deficiencies that can be repaired from the source, while recurring task demands induce broader preferences for how source knowledge should be represented. Both pathways modify the same persistent source model, and both preserve the same invariant: task experience determines what should be reconsidered, while the authoritative source determines what can become persistent knowledge.

%% file: figure/fig-overview.tex
\begin{figure}[t]
\centering
\includegraphics[width=0.98\linewidth]{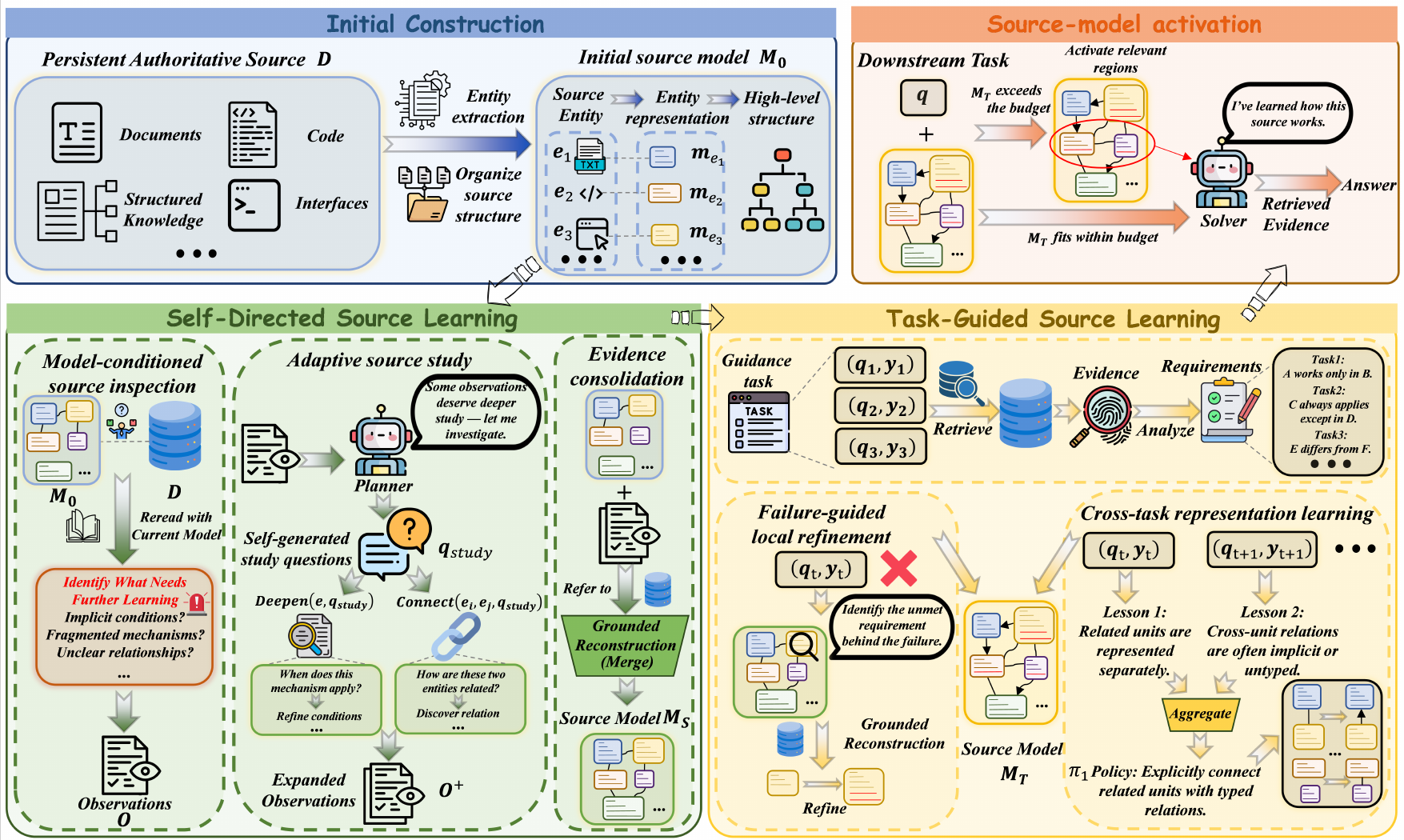}
\caption{
\textbf{Overview of \method{}.}
\textbf{Initial Construction} builds a provisional source model $M_0$ from the authoritative source $D$.
\textbf{\textit{Self-Directed Source Learning}} deepens and connects source understanding through adaptive study, producing $M_S$.
\textbf{\textit{Task-Guided Source Learning}} uses guidance-task failures and recurring representation needs to further refine the model into $M_T$.
At inference time, \textbf{Source-Model Activation} combines relevant source-model context with directly retrieved evidence for downstream tasks.
}
\label{fig:framework}
\vspace{-4mm}
\end{figure}

%% file: 5_exp.tex
We evaluate \method by addressing three questions: \textbf{Q1}: does source learning improve same-source task performance across source types and LLM backends (Section~\ref{sec:main_results})? \textbf{Q2}: how do Self-Directed and Task-Guided Source Learning contribute to the final source model (Section~\ref{sec:ablation})? \textbf{Q3}: does source learning develop task-relevant source understanding beyond static source compression (Section~\ref{sec:q3})?

\subsection{Experimental Setup}
\label{sec:exp_setup}

\noindent\textbf{Tasks and Datasets.} We evaluate \method on five benchmarks spanning document, code, and API sources: \textbf{MultiDoc2Dial}~\citep{feng2021multidoc2dial}, \textbf{NarrativeQA}~\citep{kovcisky2018narrativeqa}, and \textbf{SWE-QA}~\citep{peng2026swe} for source-grounded QA; \textbf{APIBench}~\citep{patil2024gorilla} for tool use; and \textbf{AppWorld}~\citep{trivedi2024appworld} for interactive tasks. Each document collection, story, repository, or API library defines one persistent source. Guidance and test tasks are disjoint and shared across methods; we use three fixed 30\%/70\% splits except for AppWorld, which uses its official training and test-normal splits. QA answers are judged by GPT-5.6-Luna~\citep{openai2026gpt56}, while APIBench and AppWorld use their official evaluators. Full dataset and evaluation details are provided in Appendix~\ref{app:exp_details}.

\noindent\textbf{LLM Backends.} We use GPT-5.6-Luna~\citep{openai2026gpt56}, gpt-oss-120b~\citep{agarwal2025gpt}, and DeepSeek-V4.1-Flash~\citep{xu2026deepseek}. Within each configuration, the same backend performs all method-specific LLM operations, and all methods share the same embedding model~\citep{openai2024embedding} and, on AppWorld, the same agent scaffold. At task time, \method receives the same retrieved evidence as Hybrid RAG together with the activated source model (\cref{eq:task-time-use}).

\noindent\textbf{Counterparts.} We compare against task-local retrieval with Hybrid RAG, persistent source representations with RAPTOR~\citep{sarthi2024raptor} and HippoRAG~2~\citep{gutierrez2025rag}, and experience-based memory with AWM~\citep{wang2024agent}, which learns from the same guidance tasks as \method. Implementation details for all counterparts are given in Appendix~\ref{app:counterparts}.

\subsection{Main Results}
\label{sec:main_results}
\input{table/table_main}
To answer \textbf{Q1}, Table~\ref{tab:main} reports results across all five benchmarks and three backends. \method achieves the best result in 13 of 15 settings and the second-best in one more, improving over Hybrid RAG by \textbf{+14.3}, \textbf{+4.9}, and \textbf{+13.4} points on average with GPT-5.6-Luna, gpt-oss-120b, and DeepSeek-V4.1-Flash, respectively. The gains span document, code, and API sources and exceed 10 points on four of five benchmarks with both GPT-5.6-Luna and DeepSeek-V4.1-Flash. In contrast, RAPTOR, HippoRAG~2, and AWM improve more unevenly across benchmarks, suggesting that progressively learning the source transfers more consistently than static source structuring or retaining task experience alone.

\subsection{Ablation Study}
\label{sec:ablation}
\input{figure/fig-ablation}

To address \textbf{Q2}, we remove each learning mechanism of \method{} in turn and re-evaluate on two sources of each QA benchmark with GPT-5.6-Luna, keeping the guidance tasks, test splits, and task-time source access of the main results (\cref{fig:ablation}). The left panel ablates the source learning stages. The full model $M_T$ is the most accurate on all three benchmarks. Omitting either Self-Directed or Task-Guided Source Learning lowers accuracy, and the initial source model $M_0$ is the lowest, showing that the two learning mechanisms are complementary. The right panel ablates the two pathways of Task-Guided Source Learning. Removing either failure-guided local refinement or cross-task representation learning lowers accuracy on every benchmark, and removing both, which recovers $M_S$, lowers it further. The two pathways thus carry different information from the same guidance tasks: concrete failures repair local gaps in the source model, while recurring representation lessons change how the whole model represents the source, and neither alone recovers the performance of the full model.

\subsection{Source Learning Beyond Compression}
\label{sec:q3}
\input{figure/fig-q3}

A natural concern is that the source model is merely a compressed summary of the source. To address \textbf{Q3}, we examine how its representation changes through learning and whether these changes matter for future tasks (\cref{fig:q3}).

\noindent\textbf{Learning changes what is represented.} We group source-model units into source overview, facts and attributes, rules, relations and constraints, and procedures. From $M_0$ to $M_S$ and $M_T$, the representation shifts away from isolated look-up content toward structured rules and procedures (\cref{fig:q3}a), while units with explicit applicability conditions rise from roughly 30\% in $M_0$ to 65--82\% in $M_T$ across sources (\cref{fig:q3}b). Thus, source learning reorganizes what is represented rather than merely retaining more content.

\noindent\textbf{The learned representation is more useful for future tasks.} The share of source claims required by test questions that are represented in the model rises from 23.2\% in $M_0$ to 40.0\% in $M_T$, with consistent gains across QA benchmarks (\cref{fig:q3}c). This coverage is associated with downstream success: accuracy rises from 67.5\% when none of the required claims are represented to 86.6\% when all are represented (\cref{fig:q3}d). Together, these results suggest that source learning develops a more structured and task-relevant understanding of the source, rather than merely compressing its content.

%% file: table/table_main.tex
\begin{table*}[t]
\centering
\caption{\small{
\textbf{Main results across document QA, code QA, tool use, and interactive environments.} We report task performance (\%) with changes relative to Hybrid RAG shown as subscripts. The best and second-best results are highlighted with \textbf{bold} and \underline{underline}, respectively.
}}
\vspace{-8pt}
\label{tab:main}
\scriptsize{
\resizebox{0.8\linewidth}{!}{
    \setlength\tabcolsep{2pt}
    \renewcommand\arraystretch{1.1}
    \begin{tabular}{c | c || c c | c | c | c}
    \toprule
    \noalign{\vskip -\belowrulesep}
    \rowcolor{CadetBlue!15}
    & & \multicolumn{2}{c|}{\textbf{Document QA}} & \multicolumn{1}{c|}{\textbf{Code QA}} & \multicolumn{1}{c|}{\textbf{Tool Use}} & \multicolumn{1}{c}{\textbf{Interactive Env.}} \\
    \cline{3-7}
    \rowcolor{CadetBlue!15}
    \multirow{-2}{*}{\textbf{Model}} & \multirow{-2}{*}{\textbf{Method}}
    & \textbf{MultiDoc2Dial} & \textbf{NarrativeQA} & \textbf{SWE-QA} & \textbf{APIBench} & \textbf{AppWorld} \\
    \hline
    \hline
    \multirow{5}{*}{\shortstack{GPT-5.6 \\ -Luna}}
    & \cellcolor{gray!10}Hybrid RAG & \cellcolor{gray!10}$ 64.9 $ & \cellcolor{gray!10}$ 66.0 $ & \cellcolor{gray!10}$ 56.9 $ & \cellcolor{gray!10}$ 47.8 $ & \cellcolor{gray!10}$ 72.0 $ \\
    & RAPTOR & $ \underline{71.2}\greenup{6.3} $ & $ \underline{68.1}\greenup{2.1} $ & $ 50.8\reddown{6.1} $ & $ 61.9\greenup{14.1} $ & $ 75.6\greenup{3.6} $ \\
    & \cellcolor{gray!10}HippoRAG 2 & \cellcolor{gray!10}$ 66.3\greenup{1.4} $ & \cellcolor{gray!10}$ 66.6\greenup{0.6} $ & \cellcolor{gray!10}$ \underline{62.7}\greenup{5.8} $ & \cellcolor{gray!10}$ 55.2\greenup{7.4} $ & \cellcolor{gray!10}$ \underline{77.4}\greenup{5.4} $ \\
    & AWM & $ 65.1\greenup{0.2} $ & $ 62.1\reddown{3.9} $ & $ 56.7\reddown{0.2} $ & $ \underline{62.8}\greenup{15.0} $ & $ 66.7\reddown{5.3} $ \\
    & \cellcolor{gray!10}\method & \cellcolor{gray!10}$ \mathbf{77.6\greenup{12.7}} $ & \cellcolor{gray!10}$ \mathbf{80.2\greenup{14.2}} $ & \cellcolor{gray!10}$ \mathbf{69.5\greenup{12.6}} $ & \cellcolor{gray!10}$ \mathbf{70.4\greenup{22.6}} $ & \cellcolor{gray!10}$ \mathbf{81.5\greenup{9.5}} $ \\
    \hline
    \multirow{5}{*}{\shortstack{gpt-oss \\ -120b}}
    & \cellcolor{gray!10}Hybrid RAG & \cellcolor{gray!10}$ 64.5 $ & \cellcolor{gray!10}$ 45.6 $ & \cellcolor{gray!10}$ 52.5 $ & \cellcolor{gray!10}$ 60.3 $ & \cellcolor{gray!10}$ \underline{30.4} $ \\
    & RAPTOR & $ 61.5\reddown{3.0} $ & $ 40.4\reddown{5.2} $ & $ 32.8\reddown{19.7} $ & $ 54.7\reddown{5.6} $ & $ 26.8\reddown{3.6} $ \\
    & \cellcolor{gray!10}HippoRAG 2 & \cellcolor{gray!10}$ \underline{66.7}\greenup{2.2} $ & \cellcolor{gray!10}$ \underline{45.7}\greenup{0.1} $ & \cellcolor{gray!10}$ \mathbf{56.7\greenup{4.2}} $ & \cellcolor{gray!10}$ 56.0\reddown{4.3} $ & \cellcolor{gray!10}$ \mathbf{33.9\greenup{3.5}} $ \\
    & AWM & $ 64.1\reddown{0.4} $ & $ 43.2\reddown{2.4} $ & $ 51.5\reddown{1.0} $ & $ \underline{63.2}\greenup{2.9} $ & $ 28.6\reddown{1.8} $ \\
    & \cellcolor{gray!10}\method & \cellcolor{gray!10}$ \mathbf{68.7\greenup{4.2}} $ & \cellcolor{gray!10}$ \mathbf{56.7\greenup{11.1}} $ & \cellcolor{gray!10}$ \underline{53.4}\greenup{0.9} $ & \cellcolor{gray!10}$ \mathbf{71.8\greenup{11.5}} $ & \cellcolor{gray!10}$ 27.4\reddown{3.0} $ \\
    \hline
    \multirow{5}{*}{\shortstack{DeepSeek-V4.1 \\ -Flash}}
    & \cellcolor{gray!10}Hybrid RAG & \cellcolor{gray!10}$ 63.3 $ & \cellcolor{gray!10}$ \underline{64.2} $ & \cellcolor{gray!10}$ 45.9 $ & \cellcolor{gray!10}$ 67.1 $ & \cellcolor{gray!10}$ 83.9 $ \\
    & RAPTOR & $ \underline{73.2}\greenup{9.9} $ & $ 60.4\reddown{3.8} $ & $ 36.8\reddown{9.1} $ & $ 69.3\greenup{2.2} $ & $ \underline{88.7}\greenup{4.8} $ \\
    & \cellcolor{gray!10}HippoRAG 2 & \cellcolor{gray!10}$ 67.3\greenup{4.0} $ & \cellcolor{gray!10}$ 62.0\reddown{2.2} $ & \cellcolor{gray!10}$ \underline{55.7}\greenup{9.8} $ & \cellcolor{gray!10}$ 69.6\greenup{2.5} $ & \cellcolor{gray!10}$ 86.9\greenup{3.0} $ \\
    & AWM & $ 63.7\greenup{0.4} $ & $ 61.4\reddown{2.8} $ & $ 47.4\greenup{1.5} $ & $ \underline{70.8}\greenup{3.7} $ & $ 83.9\greenup{0.0} $ \\
    & \cellcolor{gray!10}\method & \cellcolor{gray!10}$ \mathbf{81.3\greenup{18.0}} $ & \cellcolor{gray!10}$ \mathbf{76.2\greenup{12.0}} $ & \cellcolor{gray!10}$ \mathbf{61.6\greenup{15.7}} $ & \cellcolor{gray!10}$ \mathbf{82.2\greenup{15.1}} $ & \cellcolor{gray!10}$ \mathbf{89.9\greenup{6.0}} $ \\
    \noalign{\vskip -\belowrulesep}
    \bottomrule
    \end{tabular}
}}
\end{table*}

%% file: figure/fig-ablation.tex
\begin{wrapfigure}{r}{0.5\textwidth}
    \vspace{-18pt}
    \begin{center}
        \includegraphics[width=1\linewidth]{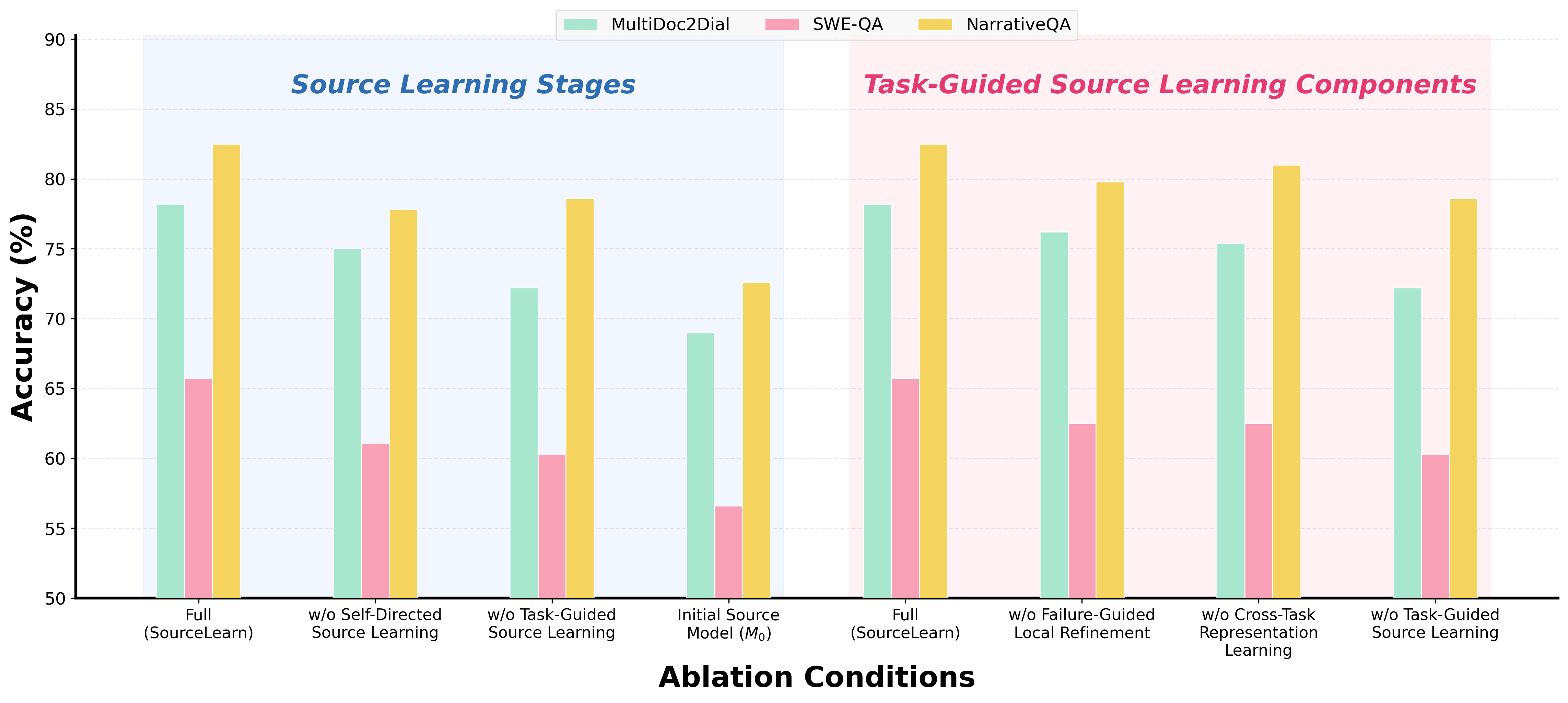}
    \end{center}
    \vspace{-10pt}
    \caption{Ablation of \method{} with GPT-5.6-Luna: the source learning stages (left) and the two pathways of Task-Guided Source Learning (right). Each bar reports answer accuracy (\%) on two sources of a QA benchmark.}
    \label{fig:ablation}
    \vspace{-10pt}
\end{wrapfigure}

%% file: figure/fig-q3.tex
\begin{figure*}[t]
    \centering
    \includegraphics[width=0.98\textwidth]{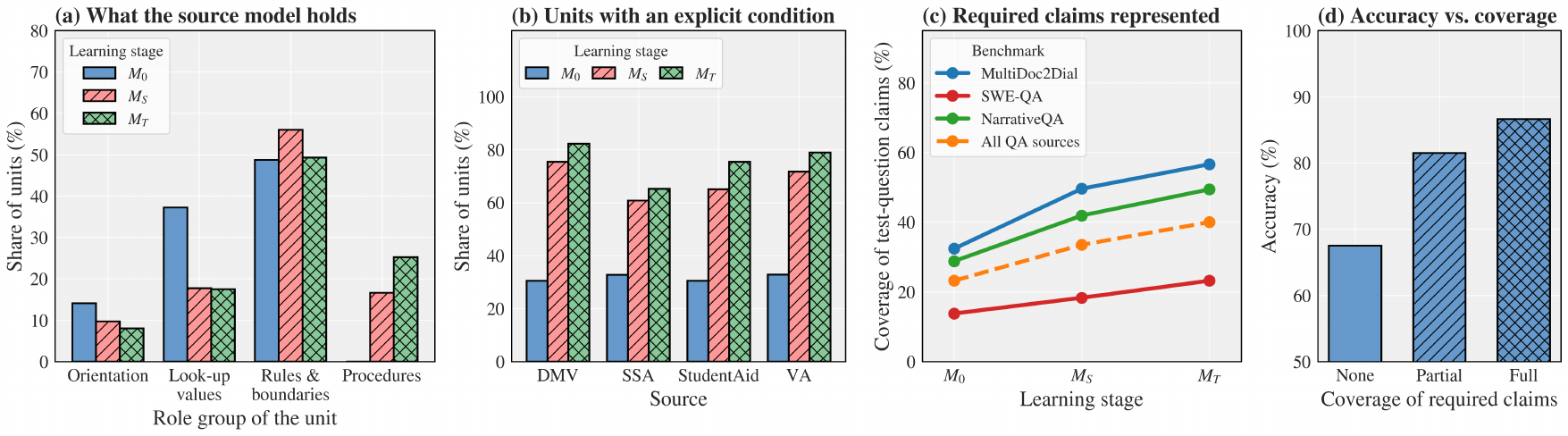}
    \vspace{-8pt}
    \caption{\small{\textbf{Source learning beyond compression.} From $M_0$ to $M_S$ to $M_T$, source learning (a) changes what the model represents, (b) makes conditions more explicit, (c) increases coverage of source knowledge required by future tasks, and (d) higher coverage is associated with higher answer accuracy.}}
    \label{fig:q3}
    \vspace{-6pt}
\end{figure*}

%% file: 6_conclusion.tex
We formulate \emph{source learning} as turning repeated interaction with a persistent authoritative source into the cumulative development of reusable \emph{source-specific competence}, rather than repeated source access or accumulation of task experience. We represent this competence with a persistent and revisable source model and propose \method{}, which develops it through complementary Self-Directed and Task-Guided Source Learning while grounding persistent updates in the original source. Across five benchmarks and three LLM backends, \method{} substantially improves overall same-source task performance over task-local retrieval, static source representations, and experience-based memory. Our analyses further show that source learning progressively makes conditions, procedures, exceptions, and relations explicit and increases coverage of the source knowledge required by future tasks, providing evidence that the learned source model goes beyond static compression. A current limitation is scope: we focus on persistent, authoritative, and relatively stable sources with repeated same-source tasks; extending source learning to evolving, noisy, or conflicting sources remains an important direction for future work.

%% file: 7_appendix.tex
\section{Experimental Details}
\label{app:exp_details}

\subsection{Dataset Details}
\label{app:exp_datasets}
Every benchmark is cast into the same form: a persistent source, a set of guidance tasks $\mathcal{G}$ used for learning, and a disjoint set of test tasks grounded in the same source. On the three QA benchmarks and APIBench, the tasks of each source are split into 30\% guidance and 70\% test tasks under three fixed seeds, keeping related tasks (questions from the same dialogue, queries for the same API) on the same side; the same splits are used by every method and backend.

\begin{itemize}[leftmargin=*]
    \item \textbf{MultiDoc2Dial}~\citep{feng2021multidoc2dial}. Goal-oriented dialogues grounded in the documents of four government domains (DMV, SSA, StudentAid, and VA). Each domain is one source (92--149 documents) with 60 questions drawn from dialogue turns.

    \item \textbf{NarrativeQA}~\citep{kovcisky2018narrativeqa}. Reading comprehension over complete stories. Each of 12 stories from the official test split is one source and is read in full rather than through the provided summaries: six movie scripts (\emph{The Breakfast Club}, \emph{Taxi Driver}, \emph{All About Eve}, \emph{Broadcast News}, \emph{Demolition Man}, and \emph{The Lord of the Rings: The Fellowship of the Ring}) and six books (\emph{Press Cuttings}, \emph{Out of Time's Abyss}, \emph{The Call of the Wild}, \emph{The Castle of Otranto}, \emph{Persuasion}, and \emph{The Old Wives' Tale}), spanning a wide range of genres and lengths, with 27--30 questions each.

    \item \textbf{SWE-QA}~\citep{peng2026swe}. Repository-level questions about the design and behavior of Python projects. We use six of the fifteen official repositories, chosen to span project kinds and sizes: streamlink (media streaming CLI), sphinx (documentation generator), xarray (labelled N-D arrays), pytest (testing framework), flask (web framework), and conan (C/C++ package manager). Each repository at its pinned commit is one source, consisting of its source code, documentation, and configuration files, with 48 questions.

    \item \textbf{APIBench}~\citep{patil2024gorilla}. Instructions that must be fulfilled by a call to the correct API of a model hub. The documentation of a hub is one source: TorchHub (186 instructions), HuggingFace (100), and TensorFlow Hub (100). Correctness is decided by the official AST-based functionality evaluator; a query whose prompt exceeds the context window of a backend is counted as a failure. The three hubs are reported as one benchmark by averaging their accuracies.

    \item \textbf{AppWorld}~\citep{trivedi2024appworld}. Everyday tasks that an agent completes by writing code against the APIs of nine simulated apps. The environment's API documentation (457 API specifications) is a single source. We use the official splits: the 90 training tasks are the guidance tasks and the 168 tasks of test-normal are the test tasks. The built-in documentation app is removed for every method, so that API documentation is reachable only through each method's own access to the source. We report the official task goal completion (TGC).
\end{itemize}

\noindent\textbf{Judging.} On the QA benchmarks, an answer is judged correct if it makes the same key claims as the reference answer without contradicting it. The judge is GPT-5.6-Luna~\citep{openai2026gpt56} at medium reasoning effort for every method, backend, and split. Accuracy is macro-averaged over the sources of a benchmark and then averaged over the three splits. APIBench and AppWorld use their official evaluators.

\subsection{Counterpart Details}
\label{app:counterparts}
All counterparts share the backend, embedding model, and answer or agent prompt with \method{}, and otherwise follow their official or reproduction settings unless noted below.

\begin{itemize}[leftmargin=*]
    \item \textbf{Hybrid RAG}. Documents are split into sections and code into files and symbol-level units. The top-30 elements by BM25~\citep{robertson2009probabilistic} and by dense retrieval~\citep{karpukhin2020dense} with text-embedding-3-large~\citep{openai2024embedding} are fused by reciprocal rank fusion~\citep{cormack2009reciprocal}, and the top 8 are passed to the solver.

    \item \textbf{RAPTOR}~\citep{sarthi2024raptor}. The official implementation with its default settings (100-token chunks, recursive clustering, cluster summaries written by the backend, collapsed-tree retrieval of the top 10 nodes within 3,500 tokens); the tree is built once per source and backend.

    \item \textbf{HippoRAG~2}~\citep{gutierrez2025rag}. The official package in its reproduction setting, with open information extraction performed by the backend; the top 8 passages are passed to the solver so that the evidence budget matches Hybrid RAG.

    \item \textbf{AWM}~\citep{wang2024agent}. The authors' offline workflow-induction pipeline. Workflows are induced by the backend from the successful guidance-task experiences of a source and split, and the induced set is placed in the prompt of every test task of that split. Source evidence at test time comes from the same Hybrid RAG retriever, so both methods receive the same task-time source evidence while differing in what reusable information they retain from the guidance tasks.
\end{itemize}

\subsection{Implementation Details}
\label{app:exp_implementation}
\noindent\textbf{Backends.} GPT-5.6-Luna and gpt-oss-120b are used at medium reasoning effort, and DeepSeek-V4.1-Flash at high effort on the QA benchmarks and medium effort on APIBench and AppWorld. Within a configuration, the same backend performs source-model construction and learning, the index construction of RAPTOR and HippoRAG~2, workflow induction for AWM, and task solving; text-embedding-3-large is the embedding model of every method.

\noindent\textbf{\method{}.} The initial source model $M_0$ is constructed once per source and backend. Self-Directed Source Learning runs one Inspect--Study--Consolidate cycle with $K=20$ adaptive study actions, and the resulting $M_S$ is shared by the three splits. Task-Guided Source Learning then processes the guidance tasks of a split in one pass, followed by recalibration under the learned representation policy. At task time, the solver receives the Hybrid RAG evidence for the task together with the activated source model under a budget of $B=24$k tokens (\cref{eq:budgeted-activation}).

\noindent\textbf{Ablation.} The ablation of Section~\ref{sec:ablation} uses two sources per QA benchmark, with the same guidance/test splits as the main results: SSA and VA from MultiDoc2Dial, pytest and conan from SWE-QA, and the scripts of \emph{All About Eve} and \emph{The Breakfast Club} from NarrativeQA. The \emph{w/o Self-Directed Source Learning} variant applies Task-Guided Source Learning directly to $M_0$; \emph{w/o Task-Guided Source Learning} corresponds to $M_S$; \emph{w/o Cross-Task Representation Learning} corresponds to $M_{\mathrm{local}}$; and \emph{w/o Failure-Guided Local Refinement} recalibrates $M_S$ under $\Pi_1$ without local refinement.

\noindent\textbf{Task solving.} On the QA benchmarks and APIBench, every method uses the same answer prompt with its own source context. On AppWorld, every method drives the official ReAct-style code agent with a limit of 50 steps; the documentation calls of the original prompt are replaced by a documentation channel that returns the method's own source context, so that methods differ only in how the API documentation is represented and accessed.

\noindent\textbf{Hardware.} GPT-5.6-Luna and DeepSeek-V4.1-Flash are accessed through their hosted APIs; gpt-oss-120b is served locally with vLLM on a server with six NVIDIA A100 80GB GPUs.

\section{Additional Results}
\label{app:additional}

All experiments in this section use GPT-5.6-Luna with the sources, guidance/test splits, and judge of the main results.

\subsection{Robustness to Incomplete Retrieval}
\label{app:degradation}
\input{figure/fig-degradation}

At task time, \method{} combines retrieved evidence $R(D,q)$ with the activated source model (\cref{eq:task-time-use}). We isolate their roles by degrading retrieval on MultiDoc2Dial, whose questions provide gold-document labels (\cref{fig:degradation}). With no retrieved evidence, the source model alone performs comparably to Hybrid RAG with all eight retrieved elements, and \method{} degrades substantially more slowly as retrieval is reduced. Removing the gold document lowers Hybrid RAG by about 25 points but \method{} by only about 8. These results show that the learned source model provides reusable source understanding when task-local retrieval is incomplete, while remaining complementary to direct access to the source.

\subsection{Comparison at Matched Context Budgets}
\label{app:token_matched}

The source model also provides more task-time context than the native retrieval counterparts. To control for this, we give each retrieval method the same per-question context budget used by \method{}, without changing its ranking or index. As shown in \cref{tab:token_matched}, \method{} remains best on three of four sources and second on Sphinx under matched context budgets. Thus, the amount of task-time context alone does not fully account for its performance; the representation analyses in Section~\ref{sec:q3} further examine what is learned beyond additional source content.

\begin{table}[H]
\centering
\caption{\small{
\textbf{Retrieval counterparts at matched context budgets} (GPT-5.6-Luna; accuracy \%). Each counterpart receives the same per-question context budget as \method{} by reading further from its own ranking. Best and second-best results are in \textbf{bold} and \underline{underlined}.
}}
\vspace{-5pt}
\label{tab:token_matched}
\scriptsize{
\resizebox{0.8\linewidth}{!}{
    \setlength\tabcolsep{6pt}
    \renewcommand\arraystretch{1.1}
    \begin{tabular}{l || c c | c | c}
    \toprule
    \noalign{\vskip -\belowrulesep}
    \rowcolor{CadetBlue!15}
    & \multicolumn{2}{c|}{\textbf{MultiDoc2Dial}} & \multicolumn{1}{c|}{\textbf{SWE-QA}} & \multicolumn{1}{c}{\textbf{APIBench}} \\
    \cline{2-5}
    \rowcolor{CadetBlue!15}
    \multirow{-2}{*}{\textbf{Method}} & \textbf{DMV} & \textbf{VA} & \textbf{Sphinx} & \textbf{HuggingFace} \\
    \hline
    \hline
    \cellcolor{gray!10}Hybrid RAG & \cellcolor{gray!10}$ \underline{76.2} $ & \cellcolor{gray!10}$ 70.2 $ & \cellcolor{gray!10}$ 62.7 $ & \cellcolor{gray!10}$ 48.6 $ \\
    RAPTOR & $ 70.2 $ & $ \underline{73.8} $ & $ 61.8 $ & $ \underline{52.9} $ \\
    \cellcolor{gray!10}HippoRAG 2 & \cellcolor{gray!10}$ 71.4 $ & \cellcolor{gray!10}$ 71.5 $ & \cellcolor{gray!10}$ \mathbf{73.5} $ & \cellcolor{gray!10}$ 51.4 $ \\
    \method & $ \mathbf{77.0} $ & $ \mathbf{77.8} $ & $ \underline{72.5} $ & $ \mathbf{56.7} $ \\
    \hline
    \cellcolor{gray!10}Evidence tokens & \cellcolor{gray!10}16.5k & \cellcolor{gray!10}18.2k & \cellcolor{gray!10}23.4k & \cellcolor{gray!10}17.1k \\
    \noalign{\vskip -\belowrulesep}
    \bottomrule
    \end{tabular}
}}
\vspace{-8pt}
\end{table}

\subsection{Grounding and Source Authority}
\label{app:grounding}

A central principle of \method{} is that learning signals determine what should be reconsidered, while persistent updates are reconstructed from the authoritative source (\cref{eq:grounded-reconstruction}). We audit this property on one document, code, and API source by judging sampled units from $M_0$, $M_S$, and $M_T$ against their cited source evidence, and by checking whether content removed during rewriting remains represented afterward (\cref{tab:grounding,tab:preservation}). Direct contradictions remain rare ($\leq 1\%$), and 93--100\% of dropped claims are preserved after rewriting. However, some Task-Guided units on the document and code sources are only partially supported or unsupported, showing that the learned representation is not itself a replacement for the source. This supports our design of keeping the original source authoritative and directly accessible while using the source model as accumulated, reusable understanding.

\begin{table}[H]
\centering
\caption{\small{
\textbf{Grounding audit of source-model units} (\%). Each unit is judged against its cited source elements (120 units per model): \emph{supported} if all claims are entailed, \emph{partial} if the main claim is supported but some detail is not, \emph{unsupported} if the cited evidence does not establish the claim, and \emph{contradicted} if it states otherwise.
}}
\vspace{-5pt}
\label{tab:grounding}
\scriptsize{
\resizebox{0.7\linewidth}{!}{
    \setlength\tabcolsep{5pt}
    \renewcommand\arraystretch{1.1}
    \begin{tabular}{l | c || c c c c}
    \toprule
    \noalign{\vskip -\belowrulesep}
    \rowcolor{CadetBlue!15}
    \textbf{Source} & \textbf{Model} & \textbf{Supported} & \textbf{Partial} & \textbf{Unsupported} & \textbf{Contradicted} \\
    \hline
    \hline
    \multirow{3}{*}{\shortstack{MultiDoc2Dial \\ (StudentAid)}}
    & \cellcolor{gray!10}$M_0$ & \cellcolor{gray!10}79.2 & \cellcolor{gray!10}19.2 & \cellcolor{gray!10}1.7 & \cellcolor{gray!10}0.0 \\
    & $M_S$ & 75.0 & 20.8 & 4.2 & 0.0 \\
    & \cellcolor{gray!10}$M_T$ & \cellcolor{gray!10}63.3 & \cellcolor{gray!10}22.5 & \cellcolor{gray!10}13.3 & \cellcolor{gray!10}0.8 \\
    \hline
    \multirow{3}{*}{\shortstack{SWE-QA \\ (pytest)}}
    & \cellcolor{gray!10}$M_0$ & \cellcolor{gray!10}89.2 & \cellcolor{gray!10}10.8 & \cellcolor{gray!10}0.0 & \cellcolor{gray!10}0.0 \\
    & $M_S$ & 86.7 & 13.3 & 0.0 & 0.0 \\
    & \cellcolor{gray!10}$M_T$ & \cellcolor{gray!10}65.0 & \cellcolor{gray!10}31.7 & \cellcolor{gray!10}3.3 & \cellcolor{gray!10}0.0 \\
    \hline
    \multirow{3}{*}{\shortstack{APIBench \\ (HuggingFace)}}
    & \cellcolor{gray!10}$M_0$ & \cellcolor{gray!10}65.0 & \cellcolor{gray!10}32.5 & \cellcolor{gray!10}1.7 & \cellcolor{gray!10}0.8 \\
    & $M_S$ & 76.7 & 18.3 & 5.0 & 0.0 \\
    & \cellcolor{gray!10}$M_T$ & \cellcolor{gray!10}84.2 & \cellcolor{gray!10}15.8 & \cellcolor{gray!10}0.0 & \cellcolor{gray!10}0.0 \\
    \noalign{\vskip -\belowrulesep}
    \bottomrule
    \end{tabular}
}}
\vspace{-8pt}
\end{table}

\begin{table}[H]
\centering
\caption{\small{
\textbf{Preservation of dropped units} (\%). Units removed by a learning stage are compared with the new model on the same entity (60 units per step). A claim is \emph{preserved} if it remains represented after rewording, merging, or condensation.
}}
\vspace{-5pt}
\label{tab:preservation}
\scriptsize{
\resizebox{0.65\linewidth}{!}{
    \setlength\tabcolsep{5pt}
    \renewcommand\arraystretch{1.1}
    \begin{tabular}{l | l || c c c}
    \toprule
    \noalign{\vskip -\belowrulesep}
    \rowcolor{CadetBlue!15}
    \textbf{Source} & \textbf{Dropped by} & \textbf{Preserved} & \textbf{Lost} & \textbf{Contradicted} \\
    \hline
    \hline
    \multirow{2}{*}{MultiDoc2Dial (StudentAid)}
    & \cellcolor{gray!10}Self-Directed ($M_0\!\to\!M_S$) & \cellcolor{gray!10}98.3 & \cellcolor{gray!10}1.7 & \cellcolor{gray!10}0.0 \\
    & Task-Guided ($M_S\!\to\!M_T$) & 98.3 & 0.0 & 1.7 \\
    \hline
    \multirow{2}{*}{SWE-QA (pytest)}
    & \cellcolor{gray!10}Self-Directed ($M_0\!\to\!M_S$) & \cellcolor{gray!10}93.3 & \cellcolor{gray!10}5.0 & \cellcolor{gray!10}1.7 \\
    & Task-Guided ($M_S\!\to\!M_T$) & 100.0 & 0.0 & 0.0 \\
    \hline
    \multirow{2}{*}{APIBench (HuggingFace)}
    & \cellcolor{gray!10}Self-Directed ($M_0\!\to\!M_S$) & \cellcolor{gray!10}96.7 & \cellcolor{gray!10}3.3 & \cellcolor{gray!10}0.0 \\
    & Task-Guided ($M_S\!\to\!M_T$) & 100.0 & 0.0 & 0.0 \\
    \noalign{\vskip -\belowrulesep}
    \bottomrule
    \end{tabular}
}}
\vspace{-8pt}
\end{table}

\subsection{Transfer Beyond Guidance-Task Regions}
\label{app:unseen}

Task-Guided Source Learning is intended to turn task experience into reusable source understanding rather than task-specific corrections. We therefore partition test questions according to whether their source region overlaps with any guidance task of the same split. As shown in \cref{tab:unseen}, \method{} improves over Hybrid RAG by 14.9 points on seen regions and 12.1 points on unseen regions. The comparable gain on unseen regions shows that the benefit extends beyond source regions directly exposed by the guidance tasks, supporting the transfer of learned source understanding to new parts of the same source.

\begin{table}[H]
\centering
\caption{\small{
\textbf{Accuracy (\%) by overlap with guidance-task regions} (GPT-5.6-Luna). \emph{Seen} regions overlap with at least one guidance task; \emph{unseen} regions do not. Subscripts show changes relative to Hybrid RAG within each stratum.
}}
\vspace{-5pt}
\label{tab:unseen}
\scriptsize{
\resizebox{0.9\linewidth}{!}{
    \setlength\tabcolsep{5pt}
    \renewcommand\arraystretch{1.1}
    \begin{tabular}{l || c c | c c | c c | c c}
    \toprule
    \noalign{\vskip -\belowrulesep}
    \rowcolor{CadetBlue!15}
    & \multicolumn{2}{c|}{\textbf{Hybrid RAG}} & \multicolumn{2}{c|}{$\boldsymbol{M_0}$} & \multicolumn{2}{c|}{$\boldsymbol{M_S}$} & \multicolumn{2}{c}{$\boldsymbol{M_T}$ \textbf{(\method)}} \\
    \cline{2-9}
    \rowcolor{CadetBlue!15}
    \multirow{-2}{*}{\textbf{Benchmark}} & Seen & Unseen & Seen & Unseen & Seen & Unseen & Seen & Unseen \\
    \hline
    \hline
    \cellcolor{gray!10}MultiDoc2Dial & \cellcolor{gray!10}$ 64.2 $ & \cellcolor{gray!10}$ 65.3 $ & \cellcolor{gray!10}$ 69.0\greenup{4.8} $ & \cellcolor{gray!10}$ 72.3\greenup{7.0} $ & \cellcolor{gray!10}$ 77.4\greenup{13.2} $ & \cellcolor{gray!10}$ 72.0\greenup{6.7} $ & \cellcolor{gray!10}$ \mathbf{78.9\greenup{14.7}} $ & \cellcolor{gray!10}$ \mathbf{76.7\greenup{11.4}} $ \\
    SWE-QA & $ 57.3 $ & $ 56.4 $ & $ 70.3\greenup{13.0} $ & $ 61.1\greenup{4.7} $ & $ 66.2\greenup{8.9} $ & $ 64.6\greenup{8.2} $ & $ \mathbf{72.9\greenup{15.6}} $ & $ \mathbf{66.3\greenup{9.9}} $ \\
    \cellcolor{gray!10}NarrativeQA & \cellcolor{gray!10}$ 65.8 $ & \cellcolor{gray!10}$ 66.2 $ & \cellcolor{gray!10}$ 74.2\greenup{8.4} $ & \cellcolor{gray!10}$ 74.0\greenup{7.8} $ & \cellcolor{gray!10}$ 77.2\greenup{11.4} $ & \cellcolor{gray!10}$ 78.4\greenup{12.2} $ & \cellcolor{gray!10}$ \mathbf{80.1\greenup{14.3}} $ & \cellcolor{gray!10}$ \mathbf{79.9\greenup{13.7}} $ \\
    \hline
    All QA & $ 62.4 $ & $ 62.9 $ & $ 71.6\greenup{9.2} $ & $ 69.6\greenup{6.7} $ & $ 73.3\greenup{10.9} $ & $ 72.2\greenup{9.3} $ & $ \mathbf{77.3\greenup{14.9}} $ & $ \mathbf{75.0\greenup{12.1}} $ \\
    \noalign{\vskip -\belowrulesep}
    \bottomrule
    \end{tabular}
}}
\vspace{-8pt}
\end{table}

\subsection{Number of Self-Directed Learning Cycles}
\label{app:cycles}

\method{} uses one Inspect--Study--Consolidate cycle by default. We extend Self-Directed Source Learning to four cycles on two MultiDoc2Dial sources to test whether repeated study provides further gains (\cref{tab:cycles}). Additional cycles continue to generate observations and expand the source model, but accuracy shows no consistent improvement beyond the first cycle. This suggests diminishing returns from repeated self-directed study, motivating the single-cycle setting used throughout the main experiments.

\begin{table}[H]
\centering
\caption{\small{
\textbf{Self-Directed Source Learning over one to four cycles} on two MultiDoc2Dial sources (GPT-5.6-Luna). Cycle 0 is $M_0$ and cycle 1 is $M_S$. Observations are inspection notes produced in each cycle; accuracy is measured over the 60 questions of each source.
}}
\vspace{-5pt}
\label{tab:cycles}
\scriptsize{
\resizebox{0.6\linewidth}{!}{
    \setlength\tabcolsep{5pt}
    \renewcommand\arraystretch{1.1}
    \begin{tabular}{c | c || c c c}
    \toprule
    \noalign{\vskip -\belowrulesep}
    \rowcolor{CadetBlue!15}
    \textbf{Source} & \textbf{Cycle} & \textbf{Units} & \textbf{Observations} & \textbf{Accuracy (\%)} \\
    \hline
    \hline
    \multirow{5}{*}{SSA}
    & \cellcolor{gray!10}0 ($M_0$) & \cellcolor{gray!10}571 & \cellcolor{gray!10}-- & \cellcolor{gray!10}73.3 \\
    & 1 ($M_S$) & 615 & 95 & 75.0 \\
    & \cellcolor{gray!10}2 & \cellcolor{gray!10}660 & \cellcolor{gray!10}53 & \cellcolor{gray!10}80.0 \\
    & 3 & 683 & 54 & 75.0 \\
    & \cellcolor{gray!10}4 & \cellcolor{gray!10}688 & \cellcolor{gray!10}45 & \cellcolor{gray!10}73.3 \\
    \hline
    \multirow{5}{*}{StudentAid}
    & \cellcolor{gray!10}0 ($M_0$) & \cellcolor{gray!10}478 & \cellcolor{gray!10}-- & \cellcolor{gray!10}73.3 \\
    & 1 ($M_S$) & 741 & 204 & 75.0 \\
    & \cellcolor{gray!10}2 & \cellcolor{gray!10}891 & \cellcolor{gray!10}149 & \cellcolor{gray!10}73.3 \\
    & 3 & 989 & 120 & 71.7 \\
    & \cellcolor{gray!10}4 & \cellcolor{gray!10}1,018 & \cellcolor{gray!10}110 & \cellcolor{gray!10}76.7 \\
    \noalign{\vskip -\belowrulesep}
    \bottomrule
    \end{tabular}
}}
\vspace{-8pt}
\end{table}
\section{Prompt Details}
\label{app:prompts}

This section presents the prompts behind the main operators of \method: initial construction ($\operatorname{Construct}$), model-conditioned inspection ($\operatorname{Observe}$), adaptive study planning ($\operatorname{PlanStudy}$), evidence consolidation with the grounding gate ($\operatorname{Consolidate}$ / $\operatorname{GroundApply}$), and cross-task representation learning ($\operatorname{Lesson}$ / $\operatorname{Aggregate}$). Prompts are quoted from the released code; every call returns a fixed JSON schema, omitted here. The remaining prompts (source-requirement diagnosis, local refinement, and the preservation check) follow the same conventions and are included in the code.

\subsection{Initial Source Model Construction ($\operatorname{Construct}$)}
Construction reads everything the source says about one entity and fills the slots of its representation $m_e$; the generic modeling instruction $\Pi_0$ states what deserves representation.
\begin{tcolorbox}[
    enhanced, breakable,
    colframe=black!70, colback=yellow!5,
    boxrule=1pt, arc=4mm,
    left=2mm, right=2mm, top=1mm, bottom=1mm,
    title={Entity-Centered Reading Prompt}
]
\small
You are reading everything the source says about ONE entity (product / module / topic) side by side, to build its profile --- not to summarize it. Fill the slots:
\begin{itemize}[leftmargin=*]
    \item \texttt{attributes}: decision-relevant properties as name/value pairs (short \texttt{snake\_case} names another entity of the same kind would also have); one element id each. Skip properties that decide nothing.
    \item \texttt{mechanisms}: how something works or a procedure's logic, when worth understanding; cite $\geq 1$ element ids.
    \item \texttt{distinctions}: things a reader must not confuse.
    \item \texttt{decisive\_details}: the few remaining facts that change a decision or behavior --- one element id each. Rarity alone is NOT a reason; it must plausibly change an answer, boundary, exception, or decision. Skip anything merely retrievable.
\end{itemize}
Prefer fewer, structural entries. \texttt{element\_id} / \texttt{element\_ids} must be copied VERBATIM from the [bracketed] ids in the material; never invent labels. The raw source stays available at use time: do NOT store facts that a reader can simply look up. Keep constraints, mechanisms, shared rules, exceptions and distinctions.

\paragraph{Generic modeling instruction $\Pi_0$}
Preserve the major entities of the source, their roles, the important relations between them, procedures, governing conditions and representative details, while avoiding redundant low-level content.
\end{tcolorbox}

\subsection{Model-Conditioned Source Inspection ($\operatorname{Observe}$)}
The same prompt drives every rereading of the source against the current model: inspection in Self-Directed Source Learning, the study actions, and recalibration under $\Pi_1$. A short note appended to it names the mode; the one for full-entity inspection is shown.
\begin{tcolorbox}[
    enhanced, breakable,
    colframe=black!70, colback=yellow!5,
    boxrule=1pt, arc=4mm,
    left=2mm, right=2mm, top=1mm, bottom=1mm,
    title={Source Inspection Prompt}
]
\small
You are studying a knowledge source to improve a compact SOURCE MODEL of it. The relevant part of the current model and the raw evidence for a learning target are shown. Record what the evidence teaches that the model does not yet explain well, as STUDY NOTES (temporary; they will be consolidated later into a rewrite of this region, so do not write model units now): each note = one observation the evidence establishes (a rule, condition, dependency, procedure step order, exception, distinction, or how two things connect), the evidence ids it rests on, its implication for the model (e.g.\ `these three facts are stages of one procedure', `unit X is missing the governing condition'), and the model units it concerns. Do not note look-up values the source answers on its own. Return an empty observations list (\texttt{learned=false}) when the model already explains the target.

\paragraph{Appended for full-entity inspection}
This is a REVIEW: the entity's complete source text is shown, so judge the model's portion about it as a whole --- is every rule, condition, procedure step, exception and cross-reference the text establishes represented, correctly scoped and not fragmented? Reconstruct only what a reader would otherwise get wrong; look-up values stay in the source.
\end{tcolorbox}

\subsection{Adaptive Source Study Planning ($\operatorname{PlanStudy}$)}
Given the inspection observations, the planner selects the next batch $\Gamma_b$ of \textsc{Deepen} and \textsc{Connect} actions, each with one focused study question (\cref{eq:study-planning}).
\begin{tcolorbox}[
    enhanced, breakable,
    colframe=black!70, colback=yellow!5,
    boxrule=1pt, arc=4mm,
    left=2mm, right=2mm, top=1mm, bottom=1mm,
    title={Study Planning Prompt}
]
\small
You direct the self-study of an agent that maintains a compact SOURCE MODEL of a knowledge source. A full inspection of the source against the model has produced the OBSERVATIONS below: temporary notes on what each entity's source establishes that the model does not yet explain well. Before the model is consolidated, choose the \texttt{\{n\}} most consequential observations to resolve, ONE observation per action: DEEPEN one entity when its observation leaves a mechanism, governing condition, step order or exception unresolved; CONNECT two entities when an observation names a relation between them the model does not express (one constraining, triggering or feeding the other; a shared rule with different instances; a procedure or dependency spanning both). For each action copy the observation it resolves and write ONE focused question --- the single thing that has to be understood to resolve that observation (for example: under what condition does Y switch to Z, and what follows?). Never ask for everything about an entity and never enumerate several aspects in one question; a question is an instrument for resolving one identified gap. Prefer entities and pairs not yet studied; never repeat an action already taken. Copy entity keys exactly as listed.
\end{tcolorbox}

\subsection{Evidence Consolidation and Grounding ($\operatorname{Consolidate}$, $\operatorname{GroundApply}$)}
Consolidation rewrites one region from its current units, the accumulated notes, and their evidence; every proposed unit then passes the grounding gate, and the rewrite is committed only if the preservation check finds no lost meaning (\cref{eq:grounded-reconstruction}).
\begin{tcolorbox}[
    enhanced, breakable,
    colframe=black!70, colback=yellow!5,
    boxrule=1pt, arc=4mm,
    left=2mm, right=2mm, top=1mm, bottom=1mm,
    title={Region Rewrite and Grounding Prompts}
]
\small
\paragraph{Region rewrite}
Revise the SOURCE MODEL of one region of a knowledge source. You are given the region's current units, the study notes gathered by inspecting and questioning this region against the raw source, and the source evidence those notes rest on. The objective: the same compact model, revised at the SAME level of abstraction so that it also carries what the notes established --- not a fuller account of the source. The source evidence is there to verify and ground what you write; it is NOT material to summarise: never restate the source passage by passage, never copy or closely paraphrase its sentences, never take over its voice (no `you', `your', `we'). Write statements ABOUT the source in the model's own register, each unit ONE reusable semantic commitment (a rule with its governing condition, a mechanism with its consequence, a procedure with its step order, a distinction, an exception to a shared rule, a dependency between entities) --- not one source excerpt, and not a replay of implementation steps unless the sequence itself is what matters. Preserve every source-supported meaning of the old units unless a stronger statement subsumes it; merge fragments that are stages of one procedure or instances of one rule; split a unit that mixes unrelated things; drop nothing the evidence supports. Never write a negative claim (something does not exist, is not supported, cannot be done) unless the source states it explicitly. Do not add look-up values the source answers on its own and do not keep isolated look-up details (a phone number, an office hour, a form number, a single value) solely for completeness. Entity cards (attribute grids) are kept as they are and are shown only as context: do not rewrite them. For each unit give its role, its scope (an entity scope copied from the old units, or the region), the old unit ids it absorbs, and the note numbers it draws on. Names, identifiers, numbers and conditions exactly as in the source; statements at most \texttt{\{chars\}} characters; declarative.

\paragraph{Grounding gate}
Decide whether the candidate statement is SUPPORTED by the source excerpts shown (entailed by them, not merely consistent). If supported, return the ids of the excerpts that establish it; if any part is not established, \texttt{supported=false} and list the unsupported parts. Source-bound rule for NEGATIVE claims: a statement saying that something does NOT exist, is NOT supported, or CANNOT be done is supported only if an excerpt explicitly states that denial; a feature merely absent from the excerpts is NOT evidence that it is absent from the source.

\paragraph{Preservation check}
A region of a source model was rewritten. Below are the OLD units' meanings (numbered) and the NEW region. Return the numbers of old meanings that the new region no longer represents --- a meaning counts as represented when the new region states it, entails it, or expresses it through a broader rule, procedure, or abstraction that covers it (merging, factoring and abstracting are fine; a lost condition, exception, value, entity or step is not).
\end{tcolorbox}

\subsection{Cross-Task Representation Learning ($\operatorname{Lesson}$, $\operatorname{Aggregate}$)}
Each guidance task yields at most one representation lesson $\ell_t$, and the lessons are aggregated into the representation policy $\Pi_1$ (\cref{eq:representation-policy}), whose rendered text is appended to every writer prompt during recalibration (\cref{eq:source-recalibration}).
\begin{tcolorbox}[
    enhanced, breakable,
    colframe=black!70, colback=yellow!5,
    boxrule=1pt, arc=4mm,
    left=2mm, right=2mm, top=1mm, bottom=1mm,
    title={Representation Lesson and Policy Aggregation Prompts}
]
\small
\paragraph{Representation lesson}
A real task was answered with a SOURCE MODEL (a compact, persistent representation of a knowledge source) next to raw retrieval. You are shown what the task required from the source (the excerpts the reference rests on and what each establishes), the model units the system had about it, and the outcome. State the ONE general representational preference this experience suggests for building a reusable model of this source: what kind of information, at what granularity, which distinction, relation or condition deserves explicit representation --- and when. A solved task shows what was used and deserves to stay represented; a failed task additionally shows what the current representation lacked or compressed away. Kinds: preserve (a type of information to keep), granularity (level of detail), distinguish (things not to conflate), relate (cross-entity relations), condition (conditions, defaults, exceptions). The lesson must contain NO source-specific identifier (class, function, field, option, file, product or value names) and must NOT restate this task's topic: name the CLASS of source content the needed knowledge belongs to (e.g.\ `the options of a configuration file', `public fields of configurable classes', `the steps of a lifecycle procedure', `exceptions to a shared rule') and what about it to represent and when, so that the same preference applies to every other instance of that class elsewhere in this source. Return lesson null when the task required nothing beyond generic reading of the source.

\paragraph{Policy aggregation}
You maintain the REPRESENTATION POLICY of a source model: a short numbered list of preferences saying what kinds of information, at what granularity, and which distinctions, relations and conditions deserve explicit representation when this knowledge source is modelled for use. You are given the current task-derived items (if any) and new lessons, one per task, each marked solved or not. Consolidate the lessons into general preferences: merge lessons that express the same preference, keep each preference actionable for a writer (what to preserve / distinguish / relate, at what level of detail, under which condition), and cite for each the lesson numbers it consolidates. A preference may restate or refine an existing item (give \texttt{keeps\_item}). No source-specific identifiers, and no single feature of the source: when several lessons concern one feature, state the preference for the CLASS of content that feature belongs to (e.g.\ `the syntax rules of any addressing or naming scheme', `the options of any configuration surface'), so that it applies to other instances in this source. At most \texttt{\{n\}} preferences, the most broadly supported first. Do not invent preferences no lesson supports.

\paragraph{Policy note appended to writer prompts}
Representation policy for this source model (what kinds of information, at what granularity and which relations deserve explicit representation; apply it when deciding what to write, keep, merge or leave to the raw source): \texttt{\{policy\}}
\end{tcolorbox}

%% file: figure/fig-degradation.tex
\begin{figure}[H]
\centering
\includegraphics[width=0.8\linewidth]{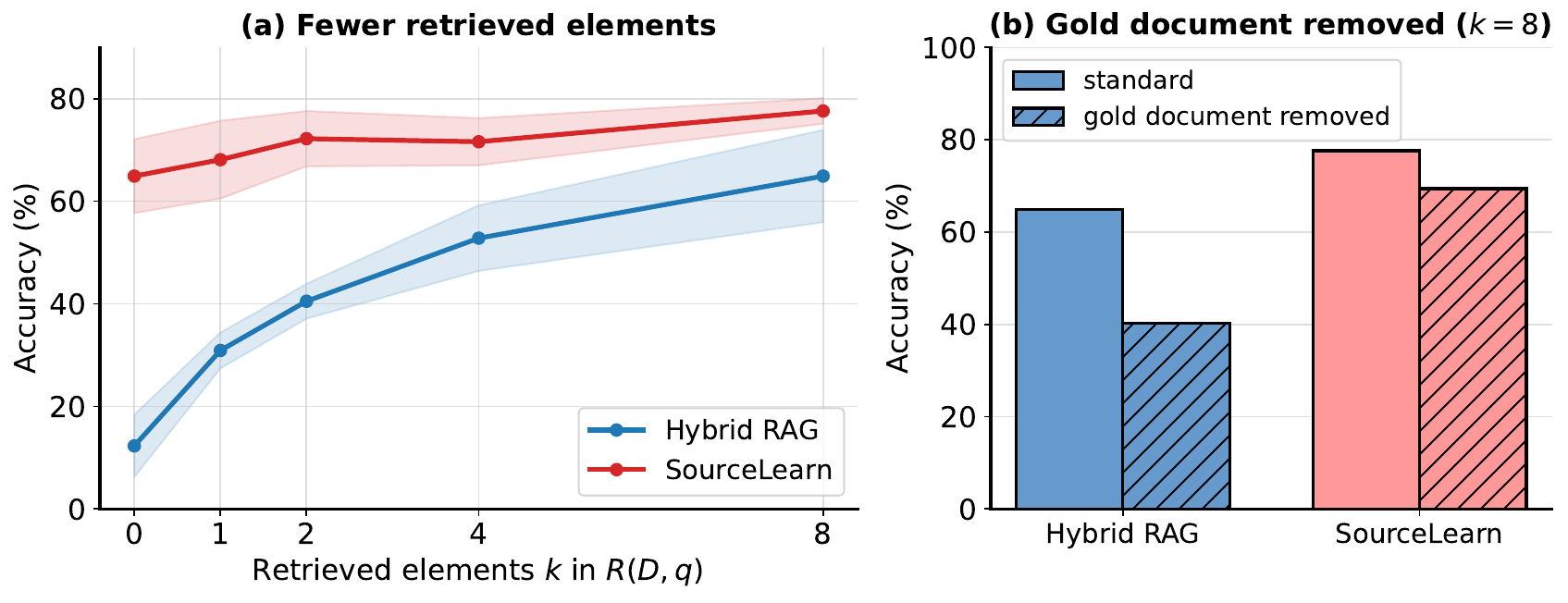}
\vspace{-6pt}
\caption{\small{
\textbf{Robustness to missing retrieved evidence} on MultiDoc2Dial with GPT-5.6-Luna.
(a) Accuracy as the number of retrieved elements decreases; at $k=0$, \method{} uses only the activated source model.
(b) Accuracy when the gold document is removed from retrieval at $k=8$.
}}
\label{fig:degradation}
\vspace{-8pt}
\end{figure}